\def\year{2022}\relax
\documentclass[letterpaper]{article} 
\makeatletter
\providecommand{\@LN}[2]{}
\makeatother
\usepackage{aaai22}  
\usepackage{times}  
\usepackage{helvet}  
\usepackage{courier}  
\usepackage[hyphens]{url}  
\usepackage{graphicx} 
\usepackage{natbib}  
\usepackage{caption} 
\DeclareCaptionStyle{ruled}{labelfont=normalfont,labelsep=colon,strut=off} 
\usepackage{algorithm}
\usepackage{algorithmic}
\usepackage{newfloat}
\usepackage{listings}
\floatstyle{ruled}
\newfloat{listing}{tb}{lst}{}
\floatname{listing}{Listing}
\newcommand{\faithfulness}{Faithfulness}
\newcommand{\confidence}{Confidence Indication}
\newcommand{\consistency}{Data Consistency}
\usepackage{comment}
\usepackage{amsmath}
\usepackage{booktabs}
\usepackage{multirow}
\usepackage{arydshln}
\usepackage{cellspace}
\usepackage{amsfonts}
\title{Diagnostics-Guided Explanation Generation}
\author{Pepa Atanasova \text{    } Jakob Grue Simonsen \text{    } Christina Lioma \text{    } Isabelle Augenstein\\
  Department of Computer Science \\
  University of Copenhagen \\
  Denmark \\
  \texttt{\{pepa, simonsen, c.lioma, augenstein\}@di.ku.dk} \\}

\begin{document}

\maketitle
\begin{abstract}
Explanations shed light on a machine learning model's rationales and can aid in identifying deficiencies in its reasoning process. Explanation generation models are typically trained in a supervised way given human explanations. When such annotations are not available, explanations are often selected as those portions of the input that maximise a downstream task's performance, which corresponds to optimising an explanation's \faithfulness\ to a given model. \faithfulness\ is one of several  so-called \textit{diagnostic properties}, which prior work has identified as useful for gauging the quality of an explanation without requiring annotations. Other diagnostic properties are \consistency, which measures how similar explanations are for similar input instances, and \confidence, which shows whether the explanation reflects the confidence of the model.
In this work, we show how to directly optimise for these diagnostic properties when training a model to generate sentence-level explanations, which markedly improves explanation quality, agreement with human rationales, and downstream task performance on three complex reasoning tasks.
\end{abstract}

\section{Introduction}
\begin{figure}[t]
\centering
\includegraphics[width=0.85\columnwidth]{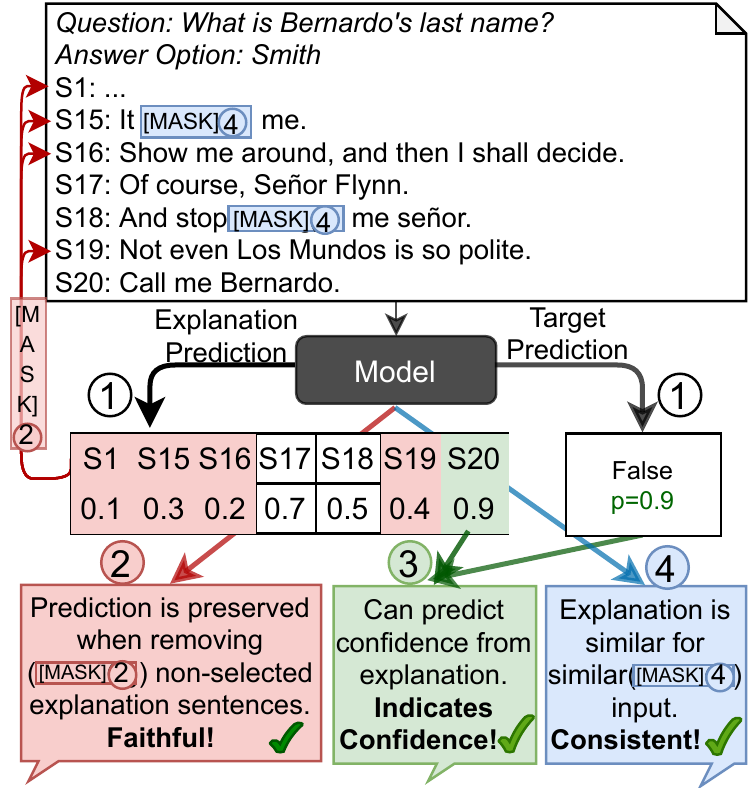}
\caption{Example instance from MultiRC with predicted target and explanation (Step 1), where sentences with confidence $\!\geq\!$ 0.5 are selected as explanations (S17, S18, S20). Steps 2-4 illustrate the use of \faithfulness, \consistency, and \confidence\ diagnostic properties as additional learning signals. `[MASK](2)' is used in Step 2 for sentences (in red) that are not explanations, and `[MASK](4)'--for random words in Step 4.}
\label{figure:example} 
\end{figure}

Explanations are an important complement to the predictions of a ML model. 
They unveil the decisions of a model that lead to a particular prediction, which increases user trust in the automated system and can help find its vulnerabilities. Moreover, ``The right $\dots$ to obtain an explanation of the decision reached'' is enshrined in the European law~\cite{regulation2016regulation}.

In NLP, 
research on explanation generation has spurred the release of datasets~\cite{zaidan-eisner-piatko-2008:nips,thorne-etal-2018-fever,khashabi-etal-2018-looking} containing human rationales for the correct predictions of downstream tasks in the form of word- or sentence-level selections of the input text. Such datasets are particularly beneficial for knowledge-intensive tasks~\cite{petroni2020kilt} with long sentence-level explanations, e.g., question answering and fact-checking, where identifying the required information is an important prerequisite for a correct prediction. They can be used to supervise and evaluate whether a model employs the correct rationales for its predictions~\cite{deyoung-etal-2020-eraser,thorne-etal-2018-fever}. The goal of this paper is to improve the sentence-level explanations generated for such complex reasoning tasks.

When human explanation annotations are not present, a common approach~\cite{lei-etal-2016-rationalizing,yu-etal-2019-rethinking} is to train models that select regions from the input maximising end task performance, which corresponds to the \textit{\faithfulness}\ property.
\citet{atanasova-etal-2020-diagnostic} propose \faithfulness\ and other \textit{diagnostic properties} to evaluate different characteristics of explanations. These include \textit{\consistency}, which measures the similarity of the explanations between similar instances, and \textit{\confidence}, which evaluates whether the explanation reflects the model's confidence, among others (see Figure \ref{figure:example} for an example).

\paragraph{Contributions} We present the first method to \textit{learn the aforementioned diagnostic properties in an unsupervised way}, directly optimising for them to improve the quality of generated explanations. We implement a joint task prediction and explanation generation model, which selects rationales at sentence level. Each property can then be included as an additional training objective in the joint model. With experiments on three complex reasoning tasks, we find that apart from improving the properties we optimised for, diagnostic-guided training also 
leads to explanations with higher agreement with human rationales, and improved downstream task performance. Moreover, we find that jointly optimising for diagnostic properties 
leads to reduced claim/question-only bias~\cite{schuster-etal-2019-towards} for the target prediction, and means that the model relies more extensively on the provided evidence.
Importantly, we also find that optimising for diagnostic properties of explanations without supervision for explanation generation does not lead to good human agreement. This indicates the need for human rationales to train models that make the right predictions for the right reasons.

\section{Related Work}
\label{sec:related}

\textbf{Supervised Explanations.} In an effort to guide ML models to perform human-like reasoning and avoid learning spurious patterns~\cite{zhang-etal-2016-rationale,ghaeini-etal-2019-saliency}, multiple datasets with explanation annotations at the word and sentence level have been proposed~\cite{wiegreffe2021teach}. 
Explanation annotations are also used for supervised explanation generation, e.g., in pipeline models, where the generation task is followed by predicting the target task from the selected rationales only ~\cite{deyoung-etal-2020-evidence,lehman-etal-2019-inferring}. As \citet{wiegreffe2020measuring,kumar-talukdar-2020-nile,jacovi2021aligning} point, pipeline models produce explanations without task-specific knowledge and without knowing the label to explain.
However, for completion, we include the baseline pipeline from ERASER's benchmark \citet{deyoung-etal-2020-evidence} as a reference model for our experiments.

Explanation generation can also be trained jointly with the target task~\cite{atanasova-etal-2020-generating-fact,li-etal-2018-end}, which has been shown to improve the performance for both tasks. Furthermore, ~\citet{wiegreffe2020measuring} suggest that self-rationalising models, such as multi-task models, provide more label-informed rationales than pipeline models. Such multi-task models can additionally learn a joint probability of the explanation and the target task prediction on the input. This can be decomposed into first extracting evidence, then predicting the class based on it~\cite{zhao2020transformer-xh,zhou-etal-2019-gear}, or vice versa~\cite{pruthi-etal-2020-weakly}. 
In this work, we also employ joint conditional training. 
It additionally provides \textit{a good testbed for our experiments with ablations of supervised and diagnostic property objectives, which is not possible with a pipeline approach.}

Most multi-task models encode each sentence separately, then combine their representations, e.g., with Graph Attention Layers~\cite{zhao2020transformer-xh,zhou-etal-2019-gear}. \citet{glockner-etal-2020-think} predict the target label from each separate sentence encoding and use the most confident sentence prediction as explanation, which also allows for unsupervised explanation generation. 
We consider \citet{glockner-etal-2020-think} as a reference model, as it is the only other work that reports results on generating explanations at the sentence level for three complex reasoning datasets from the ERASER benchmark~\cite{deyoung-etal-2020-eraser}. It also outperforms the baseline pipeline model we include from ERASER. \textit{Unlike related multi-task models, we encode the whole input jointly so that the resulting sentence representations are sensitive to the wider document context. The latter proves to be especially beneficial for explanations consisting of multiple sentences. Furthermore, while the model of \citet{glockner-etal-2020-think} is limited to a fixed and small number (up to two) of sentences per explanation, our model can predict a variable number of sentences depending on the separate instances' rationales.}


\textbf{Unsupervised Explanations.} When human explanation annotations are not provided, model's rationales can be explained with post-hoc methods based on gradients~\cite{sundararajan2017axiomatic}, simplifications~\cite{ribeiromodel}, or teacher-student setups~\cite{pruthi2020evaluating}. Another approach is to select input tokens that preserve a model's original prediction~\cite{lei-etal-2018-cooperative,yu-etal-2019-rethinking,bastings-etal-2019-interpretable,paranjape-etal-2020-information}, which corresponds to the \faithfulness\ property of an explanation. However, as such explanations are not supervised by human rationales, they do not have high overlap with human annotations~\cite{deyoung-etal-2020-eraser}. Rather, they explain what a model has learned, which does not always correspond to correct rationales and can contain spurious patterns~\cite{wang-culotta-2020-identifying}. 


\section{Method}
\label{sec:method}
We propose a novel Transformer~\cite{vaswani2017attention} based model to jointly optimise sentence-level explanation generation and downstream task performance. The joint training provides a suitable testbed for our experiments with supervised and diagnostic property objectives for a single model. The joint training optimises two training objectives for the two tasks at the same time. By leveraging information from each task, the model is guided to predict the target task based on correct rationales and to generate explanations based on the model's information needs for target prediction. This provides additional useful information for training each of the tasks. Conducting joint training for these two tasks was shown to improve the performance for each of them~\cite{zhao2020transformer-xh,atanasova-etal-2020-generating-fact}. 

The \textbf{core novelty} is that the model is trained to improve the quality of its explanations by using diagnostic properties of explanations as additional training signals (see Figure~\ref{figure:example}). We select the properties \faithfulness, \consistency, and \confidence, as they can be effectively formulated as training objectives. \faithfulness\ is also employed in explainability benchmarks~\cite{deyoung-etal-2020-eraser} and in related work for unsupervised token-level explanation generation~\cite{lei-etal-2016-rationalizing,lei-etal-2018-cooperative}, whereas we consider it at sentence level. Further, multiple studies~\cite{NEURIPS2019_a7471fdc,alvarez2018robustness} find that explainability techniques are not robust to insignificant and/or adversarial input perturbations, which we address with the \consistency\ property. We do not consider Human Agreement and Rationale Consistency, proposed in \citet{atanasova-etal-2020-diagnostic}. The supervised explanation generation training employs human rationale annotations and thus addresses Human Agreement. Rationale Consistency requires the training of a second model, which is resource-expensive.  Another property to investigate in future work is whether a model's prediction can be simulated by another model trained only on the explanations~\cite{hase-etal-2020-leakage,treviso-martins-2020-explanation,pruthi-etal-2020-learning}, which also requires training an additional model. We now describe each component in detail.

\subsection{Joint Modelling}
Let $D\!=\!\{(x_i, y_i, e_i)|i\in[1, \#(D)]\}$ be a classification dataset. The textual input $x_i\!=\!(q_i, a_i^{[opt]}, s_{i})$ consists of a question or a claim, an optional answer, and several sentences (usually above 10) $s_i\!=\!\{s_{i,j}|j\!\in\![1, \#(s_i)]\}$ used for predicting a classification label $y_i\!\in\![1, N]$. Additionally, D contains human rationale annotations selected from the sentences $s_i$ as a binary vector $e_i\!=\!\{{e_{i,j}\!=\!\{0,1\}|j\!\in\![1, \#(s_i)]}\}$, which defines a binary classification task for explanation extraction.

First, the joint model takes $x_i$ as input and encodes it using a Transformer model, resulting in contextual token representations $h^L = \textit{encode}(x_i)$ from the final Transformer layer $L$. From $h^L$, we select the representations of the CLS token that precedes the question--as it is commonly used for downstream task prediction in Transformer architectures--and the CLS token representations preceding each sentence in $s_i$, which we use for selecting sentences as an explanation. 
The selected representations are then transformed with two separate linear layers - $h^C$ for predicting the target, and $h^E$ for generating the explanations, which have the same hidden size as the size of the contextual representations in $h^L$. 

Given representations from $h^E$, a N-dimensional linear layer predicts the importance $p^E \! \in\! \mathbb{R}^{\#(s_i)}$ of the evidence sentences for the prediction of each class. As a final sentence importance score, we only take the score for the predicted class $p^{E[c]}$ and add a sigmoid layer on top for predicting the binary explanation selection task. Given representations from $h^C$, a N-dimensional linear layer with a soft-max layer on top predicts the target label $p^{C'}\!\in\!\mathbb{R}$. The model then predicts the joint conditional likelihood $L$ of the target task and the generated explanation given the input (Eq.~\ref{eq:joint}). This is factorised further into first extracting the explanations conditioned on the input and then predicting the target label (Eq.~\ref{eq:factor}) based on the extracted explanations (assuming  $\left.y_{i} \perp \mathbf{x}_{i} \mid \mathbf{e}_{i}\right)$).
\begin{gather}
L=\prod_{i=1}^{\#(D)} p\left(y_{i}, \mathbf{e}_{i} \mid \mathbf{x}_{i}\right) ~\label{eq:joint} \\
L =\prod_{i=1}^{\#(D)} p\left(\mathbf{e}_{i} \mid \mathbf{x}_{i}\right) p\left(y_{i} \mid \mathbf{e}_{i}\right)~\label{eq:factor}
\end{gather}
We condition the label prediction on the explanation prediction by multiplying $p^{C'}$ and $p^E$, resulting in the final prediction $p^{C}\! \in\! \mathbb{R}$. The model is trained to optimise jointly the target task cross-entropy loss function ($\mathcal{L}_C$) and the explanation generation cross-entropy loss function ($\mathcal{L}_E$):
\begin{gather}
\mathcal{L} = \mathcal{L}_C(p^C, y) + \mathcal{L}_E(p^{E[c]}, e)
\end{gather}
All loss terms of the diagnostic explainability properties described below are added to $\mathcal{L}$ without additional hyper-parameter weights for the separate loss terms.

\subsection{\faithfulness\ (F)}
The \faithfulness\ property guides explanation generation to select sentences preserving the original prediction, (Step 2,  Fig.~\ref{figure:example}). 
In more detail, we take sentence explanation scores $p^{E[c]}\!\in\![0,1]$ and sample from a Bernoulli distribution the sentences which should be preserved in the input: $c^{E}\!\sim\! \textit{Bern}(p^{E[c]})$. Further, we make two predictions -- one, where only the selected sentences are used as an input for the model, thus producing a new target label prediction $l^{S}$, and one where we use only unselected sentences, producing the new target label prediction $l^{Co}$. The assumption is that a high number $\#(l^{C=S})$ of predictions $l^{S}$ matching the original $l^C$ indicate the sufficiency (S) of the selected explanation. On the contrary, a low number $\#(l^{C=Co})$  of predictions $l^{Co}$ matching the original $l^C$ indicate the selected explanation is complete (Co) and no sentences indicating the correct label are missed. We then use the REINFORCE~\cite{williams1992simple} algorithm to maximise the reward:
\begin{gather}
\begin{split}\mathcal{R}_F = {\scriptstyle \#}(l^{C=S}) - {\scriptstyle \#}(l^{C=Co})- |{\scriptstyle \%}(c^E) - \lambda|
\end{split}
\end{gather}
The last term is an additional sparsity penalty for selecting more/less than $\lambda$\% of the input sentences as an explanation, $\lambda$ is a hyper-parameter.

\subsection{\consistency\ (DC)}
\consistency\ measures how similar the explanations for similar instances are. Including it as an additional training objective can serve as regularisation for the model to be consistent in the generated explanations. To do so, we mask $K$ random words in the input, where $K$ is a hyper-parameter depending on the dataset. We use the masked text (M) as an input for the joint model, which predicts new sentence scores $p^{EM}$. We then construct an $\mathcal{L}1$ loss term for the property to minimise for the absolute difference between $p^{E}$ and $p^{EM}$:
\begin{gather}
\mathcal{L}_{DC} = |p^{E} - p^{EM}|
\end{gather}
\noindent We use $\mathcal{L}1$ instead of $\mathcal{L}2$ loss as we do not want to penalise for potentially masking important words, which would result in entirely different outlier predictions. 

\subsection{\confidence\ (CI)}
The \confidence\ property measures whether generated explanations reflect the confidence of the model's predictions (Step 3, Fig.~\ref{figure:example}). We consider this a useful training objective to re-calibrate and align the prediction confidence values of both tasks.
To learn explanations that indicate prediction confidence, we aggregate the sentence importance scores, taking their maximum, minimum, mean, and standard deviation. We transform the four statistics with a linear layer that predicts the confidence $\hat{p}^C$ of the original prediction. We train the model to minimise $\mathcal{L}1$ loss between $\hat{p}^C$ and $p^C$:
\begin{gather}
\mathcal{L}_{CI} = |p^C - \hat{p}^C|
\end{gather}
\noindent We choose $\mathcal{L}1$ as opposed to $\mathcal{L}2$ loss as we do not want to penalise possible outliers due to sentences having high confidence for the opposite class.

\section{Experiments}
\begin{table*}[t]
\centering
\fontsize{10}{10}\selectfont
\begin{tabular}{llll|lll|l}
\toprule
\bf Dataset & \bf Method & \bf F1-C & \bf Acc-C & \bf P-E & \bf R-E & \bf F1-E & \bf Acc-Joint \\
\midrule
\multirow{8}{*}{\bf FEVER} & Blackbox{\tiny \cite{glockner-etal-2020-think}}& 90.2 {\scriptsize $\pm$0.4} & 90.2 {\scriptsize $\pm$0.4} & & & & \\
& Pipeline{\tiny ~\cite{deyoung-etal-2020-eraser}} & 87.7 & 87.8 & 88.3 & 87.7 & 88.0 & 78.1\\
& Supervised {\tiny \cite{glockner-etal-2020-think}} & 90.7  {\scriptsize $\pm$0.7} & 90.7  {\scriptsize $\pm$0.7} & 92.3  {\scriptsize $\pm$0.1} & 91.6  {\scriptsize $\pm$0.1} & 91.9  {\scriptsize $\pm$0.1} & 83.9  {\scriptsize $\pm$0.1}\\
\cdashline{2-8}[2pt/2pt]
& Supervised & 89.3 {\scriptsize $\pm$0.4}& 89.4 {\scriptsize $\pm$0.3} & 94.0 {\scriptsize $\pm$0.1} & 93.8 {\scriptsize $\pm$0.1} & 93.9 {\scriptsize $\pm$0.1} & 80.1 {\scriptsize $\pm$0.4} \\
& Supervised+\consistency & \bf 89.7 {\scriptsize $\pm$0.5} & \bf 89.7 {\scriptsize $\pm$0.5} &\bf 94.4 {\scriptsize $\pm$0.0} & \bf 94.2 {\scriptsize $\pm$0.0} & \bf 94.4 {\scriptsize $\pm$0.0} & 80.8 {\scriptsize $\pm$0.5}\\
& Supervised+\faithfulness & 89.5 {\scriptsize $\pm$0.4} & 89.6 {\scriptsize $\pm$0.4}  & 92.8 {\scriptsize $\pm$0.2}& 93.7 {\scriptsize $\pm$0.2}& 93.3 {\scriptsize $\pm$0.2}& 75.4 {\scriptsize $\pm$0.3}\\
& Supervised+\confidence & 87.9 {\scriptsize $\pm$1.0} & 87.9 {\scriptsize $\pm$1.0} & 93.9 {\scriptsize $\pm$0.1} & 93.7 {\scriptsize $\pm$0.1} & 93.8 {\scriptsize $\pm$0.1} & 78.5 {\scriptsize $\pm$0.9}\\
& Supervised+All &89.6 {\scriptsize $\pm$0.1}& 89.6 {\scriptsize $\pm$0.1} & 94.4 {\scriptsize $\pm$0.1} & 94.2 {\scriptsize $\pm$0.1}& 94.3 {\scriptsize $\pm$0.1} & \bf 80.9 {\scriptsize $\pm$0.1}\\
\midrule    
\multirow{8}{*}{\bf MultiRC} & Blackbox{\tiny \cite{glockner-etal-2020-think}} & 67.3 {\scriptsize $\pm$1.3} & 67.7 {\scriptsize $\pm$1.6} & & & & \\
& Pipeline{\tiny ~\cite{deyoung-etal-2020-eraser}} & 63.3 & 65.0 & 66.7 & 30.2 & 41.6 & 0.0 \\
& Supervised {\tiny \cite{glockner-etal-2020-think}} & 65.5  {\scriptsize $\pm$3.6}& 67.7  {\scriptsize $\pm$1.5} & 65.8  {\scriptsize $\pm$0.2} & 42.3  {\scriptsize $\pm$3.9} & 51.4  {\scriptsize $\pm$2.8}& 7.1  {\scriptsize $\pm$2.6}\\ 
\cdashline{2-8}[2pt/2pt]
& Supervised & 71.0 {\scriptsize $\pm$0.3}& 71.4 {\scriptsize $\pm$0.3}& 78.0 {\scriptsize $\pm$0.1}& 78.6 {\scriptsize $\pm$0.5}& 78.3 {\scriptsize $\pm$0.1}& 16.2 {\scriptsize $\pm$0.4}\\
& Supervised+\consistency & \bf 71.7 {\scriptsize $\pm$0.6} & \bf 72.2 {\scriptsize $\pm$0.7}& \bf 79.9 {\scriptsize $\pm$0.4}& 79.0 {\scriptsize $\pm$0.8}& 79.4 {\scriptsize $\pm$0.5}& \bf 19.3 {\scriptsize $\pm$0.4}\\
& Supervised+\faithfulness & 71.0 {\scriptsize $\pm$0.4}& 71.3 {\scriptsize $\pm$0.4}& 78.2 {\scriptsize $\pm$0.1} & 79.1 {\scriptsize $\pm$0.2}& 78.6 {\scriptsize $\pm$0.1} & 16.1 {\scriptsize $\pm$0.5}\\
& Supervised+\confidence & 70.6 {\scriptsize $\pm$0.7}& 71.1 {\scriptsize $\pm$0.6}& 77.9 {\scriptsize $\pm$0.8}& 78.3 {\scriptsize $\pm$0.5}& 78.1  {\scriptsize $\pm$0.5}& 16.5  {\scriptsize $\pm$1.0}\\
& Supervised+All & 70.5 {\scriptsize $\pm$1.6} & 71.2 {\scriptsize $\pm$1.3}& 79.7 {\scriptsize $\pm$1.1}& \bf 79.4 {\scriptsize $\pm$0.5}& \bf 79.6 {\scriptsize $\pm$0.7}& 18.8 {\scriptsize $\pm$1.6}\\
\midrule
\multirow{8}{*}{\bf Movies} & Blackbox{\tiny \cite{glockner-etal-2020-think}} & 90.1 {\scriptsize $\pm$0.3} & 90.1 {\scriptsize $\pm$0.3}  & & & & \\
& Pipeline{\tiny ~\cite{deyoung-etal-2020-eraser}}  & 86.0 & 86.0 & 87.9 & 60.5 & 71.7 & 40.7 \\
& Supervised {\tiny \cite{glockner-etal-2020-think}} & 85.6 {\scriptsize $\pm$3.6} & 85.8 {\scriptsize$\pm$3.5} & 86.9 {\scriptsize $\pm$2.5} & 62.4 {\scriptsize $\pm$0.1} & 72.6 {\scriptsize $\pm$0.9} & 43.9 {\scriptsize $\pm$0.6}\\ 
\cdashline{2-8}[2pt/2pt]
& Supervised & 87.4 {\scriptsize $\pm$0.4} & 87.4 {\scriptsize $\pm$0.4} & 79.6 {\scriptsize $\pm$0.6} & 68.9 {\scriptsize $\pm$0.5} & 73.8 {\scriptsize $\pm$0.5} & 59.4 {\scriptsize $\pm$0.6}\\
& Supervised+\consistency & \bf 90.0 {\scriptsize $\pm$0.7}& \bf 90.0 {\scriptsize $\pm$0.7}& 79.5 {\scriptsize $\pm$0.1} & 69.2 {\scriptsize $\pm$0.7} & 74.0 {\scriptsize $\pm$0.8} & 60.8 {\scriptsize $\pm$1.7}\\
& Supervised+\faithfulness & 89.1 {\scriptsize $\pm$0.6}& 89.1 {\scriptsize $\pm$0.6} & \bf 80.9 {\scriptsize $\pm$0.9}& \bf 69.9 {\scriptsize $\pm$1.3} & \bf 74.9 {\scriptsize $\pm$1.1} & \bf 62.6 {\scriptsize $\pm$1.6}\\
& Supervised+\confidence & 89.9 {\scriptsize $\pm$0.7}&89.9 {\scriptsize $\pm$0.7}& 79.7 {\scriptsize $\pm$1.4}& 69.5 {\scriptsize $\pm$0.7}& 74.3 {\scriptsize $\pm$1.0} & 60.1 {\scriptsize $\pm$2.6}\\
& Supervised+All & 89.9 {\scriptsize $\pm$0.7} & 89.9 {\scriptsize $\pm$0.7} & 80.0 {\scriptsize $\pm$1.0}& 69.5 {\scriptsize $\pm$1.0} & 74.4 {\scriptsize $\pm$1.0} & 60.3 {\scriptsize $\pm$2.2}\\
\bottomrule
\end{tabular}
\caption{Target task prediction (F1-C, Accuracy-C) and explanation generation (Precision-E, Recall-E, F1-E) results (mean and standard deviation over three random seed runs). Last columns measures joint prediction of target accuracy and explanation generation. The property with the \textbf{best relative improvement over the supervised model is in bold}.}
\label{tab:results:supervised}
\end{table*}
\label{sec:experiments}

\subsection{Datasets} 
We perform experiments on three datasets from the ERASER benchmark~\cite{deyoung-etal-2020-eraser} (FEVER, MultiRC, Movies), all of which require complex reasoning and have sentence-level rationales. For FEVER~\cite{thorne-etal-2018-fever}, given a claim and an evidence document, a model has to predict the veracity of a claim$\in$\{support, refute\}. 
The evidence for predicting the veracity has to be extracted as explanation. For MultiRC~\cite{khashabi-etal-2018-looking}, given a question, an answer option, and a document, a model has to predict if the answer is correct. For Movies~\cite{zaidan-eisner-piatko-2008:nips}, the sentiment$\in$\{positive, negative\} of a long movie review has to be predicted. For Movies, as in \citet{glockner-etal-2020-think}, we mark each sentence containing annotated explanation at token level as an explanation. 

Note that, in knowledge-intensive tasks such as fact checking and question answering also explored here, human rationales point to regions in the text containing the information needed for prediction. 
Identifying the required information becomes an important preliminary for the correct prediction rather than a plausibility indicator~\cite{jacovi-goldberg-2020-towards}, and is evaluated as well (e.g., FEVER score, Joint Accuracy) (see further discussion in supplemental material).

\subsection{Metrics} 
We evaluate the effect of using diagnostic properties as additional training objectives for explanation generation. We first measure their effect on selecting human-like explanations by evaluating precision, recall, and macro F1-score against human explanation annotations provided in each dataset (Sec. \ref{sec:explanation_results}). Second, we compute how generating improved explanations affects the target task performance by computing accuracy and macro F1-score for the target task labels (Sec. \ref{sec:target_results}). Additionally, as identifying the required information in knowledge-intensive datasets, such as FEVER and MultiRC, is an important preliminary for a correct prediction, and following \citet{thorne-etal-2018-fever, glockner-etal-2020-think}, we evaluate the joint target and explanation performance by considering a prediction as correct only when the whole explanation is retrieved (Acc. Full). In case of multiple possible explanations $e_i$ for one instance (ERASER provides comprehensive explanation annotations for the test sets), selecting one of them counts as a correct prediction. Finally, as diagnostic property training objectives target particular properties, we measure the improvements for each property (Sec. \ref{sec:properties_results}).

\subsection{Experimental Setting} 
Our core goal is to measure the relative improvement of the explanations generated by the underlying model with (as opposed to without) diagnostic properties. We conduct experiments for the supervised model (Sup.), including separately \faithfulness\ (F), \consistency\ (DC), and \confidence\ (CI), as well as all three (All) as additional training signals (see Section~\ref{sec:method}).

Nevertheless, we include results from two other architectures generating sentence-level explanations that serve as a reference for explanation generation performance on the employed datasets. Particularly, we include the best supervised sentence explanation generation results reported in \citet{glockner-etal-2020-think}, and the baseline pipeline model from ERASER~\cite{deyoung-etal-2020-evidence}, which extracts one sentence as explanation and uses it only for target prediction (see Sec.~\ref{sec:related} for a detailed comparison). We also include an additional baseline comparison for the target prediction task. The BERT Blackbox model predicts the target task from the whole document as an input without being supervised by human rationales. The results are as reported by \citet{glockner-etal-2020-think}. In our experiments, we also use BERT~\cite{devlin-etal-2019-bert} base-uncased as our base architecture, following \citet{glockner-etal-2020-think}.


\section{Results}
\subsection{Explanation Generation Results}
~\label{sec:explanation_results}
In Table~\ref{tab:results:supervised}, we
see that our supervised model performs better than
~\citet{glockner-etal-2020-think,deyoung-etal-2020-evidence}.
For the MultiRC dataset, where the explanation consists of more than one sentence, our model brings an improvement of more than 30 F1 points over the reference models, confirming the importance of the contextual information, which performs better than encoding each explanation sentence separately.

When using the diagnostic properties as additional training objectives, we see further improvements in the generated explanations. The most significant improvement is achieved with the \consistency\ property for all datasets with up to 2.5 F1 points over the underlying supervised model. We assume that the \consistency\ objective can be considered as a regularisation for the model's instabilities at the explanation level. The second highest improvement is achieved with the \faithfulness\ property, increasing F1 by up to 1 F1 point for Movies and MultiRC. We assume that the property does not result in improvements for FEVER as it has multiple possible explanation annotations for one instance, which can make the task of selecting one sentence as a complete explanation ambiguous. \confidence\ results in improvements only on Movies. We conjecture that \confidence\ is the least related to promoting similarity to human rationales in the generated explanations. Moreover, the re-calibration of the prediction confidence for both tasks possibly leads to fewer prediction changes, explaining the low scores w.r.t. human annotations. We look into how \confidence\ affects the selected annotations in Sections~\ref{sec:properties_results}, and~\ref{sec:discussion}. Finally, combining all diagnostic property objectives, results in a performance close to the best performing property for each dataset.

\subsection{Target Prediction Results}
~\label{sec:target_results}
In Table~\ref{tab:results:supervised}, the Supervised model, without additional property objectives, consistently improves target task performance by up to 4 points in F1, compared to the two reference models that also generate explanations, except for the FEVER dataset, where the models already achieve high results. This can be due to the model encoding all explanation sentences at once, which allows for a more informed prediction of the correct target class. Our model trained jointly with the target task and explanation prediction objective also has similar performance to the BERT Blackbox model and even outperforms it by 4.4 F1 points for the MultiRC dataset. Apart from achieving high target prediction performance (F1-C) on the target task, our supervised model also learns which parts of the input are most important for the prediction, which is an important prerequisite for knowledge-intensive tasks. 

We see further improvements in downstream task performance when using the diagnostic properties as additional training objectives. Improvements of the generated explanations usually lead to improved target prediction as they are conditioned on the extracted evidence. Here, we again see that \consistency\ steadily improves the target task's performance with up to 2.5 F1 points. We also see improvements in F1 with \faithfulness\ for FEVER and MultiRC. Finally, we find that improvements in \confidence\ lead to an improvement for target prediction of 2.5 F1 points for Movies. Combining all objectives, results in performance close the performance of the other properties.

We also show joint prediction results for target task and evidence. For MultiRC and Movies, the improvements of our supervised model over~\citet{glockner-etal-2020-think} are very considerable with up to 9 accuracy points; using diagnostic properties increases results further up to 4 points in accuracy. Apart from improving the properties of the generated explanations, this could be due to the architecture conditioning the prediction on the explanation. The only dataset we do not see improvements for is FEVER, where again the performance is already high, and the target prediction of our model performs worse than \citet{glockner-etal-2020-think}.

\subsection{Explanations Property Results}
~\label{sec:properties_results}
So far, we concentrate on the relative performance improvements compared to human annotations. However, the diagnostic properties' additional training objectives are directed at generating explanations that exhibit these properties to a larger degree. Here, we demonstrate the improvements over the explanation properties themselves for unseen instances in the test splits. Note that this is a control experiment as we expect the properties we optimise for to be improved.

\begin{table}[t]
\centering
\fontsize{10}{10}\selectfont
\begin{tabular}{llrr}
\toprule
\textbf{Dataset} & \textbf{Method} & \textbf{Suff. $\uparrow$} & \textbf{Compl. $\downarrow$}\\
\midrule
\multirow{2}{*}{\bf FEVER}  & Supervised & 85.1 & 85.1 \\ 
& Supervised+F & 97.4 & 83.6 \\
\midrule
\multirow{2}{*}{\bf MultiRC}  & Supervised & 81.7 & 69.2\\
& Supervised+F & 82.3 & 67.0\\ \midrule
\multirow{2}{*}{\bf Movies}  & Supervised & 94.8 & 92.2\\
& Supervised+F & 96.6 & 91.3\\
\bottomrule
\end{tabular}
\caption{Sufficiency and Completeness as proportions of the instances that preserve their prediction when evaluated on only the selected (Suff.) or the unselected (Compl.) explanation sentences, accordingly, for training with and without the \faithfulness\ objective.}~\label{tab:results:faith}
\end{table}

\textbf{\faithfulness.} In Table~\ref{tab:results:faith}  we see that supervision from the \faithfulness\ property leads to generating explanations that preserve the original label of the instance for all datasets. For FEVER, the label is even preserved in 12\% of the instances more than with the supervised objective only. The least faithful explanations are those generated for MultiRC, which can be explained by the low joint performance of both tasks. We also see that even when removing the selected explanations, it is still possible to predict the same label based on the remaining evidence. Such cases are decreased when including the \faithfulness\ property. The latter phenomenon can be explained by the fact that FEVER and Movies' instances contain several possible explanations. We conjecture that this might also be due to the model learning spurious correlations. We further study this in Section~\ref{sec:bias}.

\begin{table}[t]
\centering
\fontsize{10}{10}\selectfont
\begin{tabular}{llrr}
\toprule
\textbf{Dataset} &\textbf{Method} & \textbf{Pred.} & \textbf{Expl.}\\
\midrule
\multirow{2}{*}{\bf FEVER} & Sup. & 0.03 (9.9e-8) & 3.68 (1.80)\\
 & Sup.+DC & 0.02 (9.1e-8) & 2.56 (0.97)\\ \midrule
\multirow{2}{*}{\bf MultiRC} & Sup. & 0.09 (5.6e-8) & 7.83(2.87)\\
 & Sup.+DC & 0.05 (4.9e-8) & 3.01(0.89)\\ \midrule
\multirow{2}{*}{\bf Movies} & Sup. & 0.04 (7.1e-8)& 2.34 (1.38)\\
 & Sup.+DC & 0.01 (6.2e-8) & 1.72 (0.90)\\
\bottomrule
\end{tabular}
\caption{Mean and standard deviation (in brackets) of the difference between target (Pred.) and explanation (Expl.) prediction confidence for similar (masked) instances.}
\label{tab:results:stability}
\end{table}
\begin{table}[t]
\centering
\fontsize{10}{10}\selectfont
\begin{tabular}{lrrr}
\toprule
\textbf{Method} & \textbf{FEVER} & \textbf{MultiRC} & \textbf{Movies}\\
\midrule
Sup. & 0.10 (0.17) & 0.05 (0.10) & 0.12 (0.09)\\
Sup.+CI & 0.05 (0.09) & 0.04 (0.09) & 0.05 (0.10) \\
\bottomrule
\end{tabular}
\caption{Mean and standard deviation (in brackets) difference between the model's confidence and the confidence of the generated explanations.}
\label{tab:results:confidence}
\end{table}

\textbf{\consistency.} Using \consistency\ as an additional training objective aims to regularise the model to select similar explanations for similar instances. In Table~\ref{tab:results:stability}, we find the variance of downstream task prediction confidence decreases for all datasets with up to 0.04 points. Furthermore, the variance of generated explanation probabilities for similar instances is decreased as well. The largest improvements are for MultiRC and Movies, where the property brings the highest performance improvement w.r.t. human annotations as well. We also find that the Movies dataset, which has the longest inputs, has the smallest variance in explanation predictions. This suggests that the variance in explanation prediction is more pronounced for shorter inputs as in FEVER and MultiRC, where the property brings more improvement w.r.t. human annotations. The variance could also depend on the dataset's nature.

\textbf{\confidence.} Table~\ref{tab:results:confidence} shows the difference between the confidence of the predicted target label and the confidence of the explanation sentence with the highest importance. Including \confidence\ as a training objective indeed decreases the distance between the confidence of the two tasks, making it easier to judge the confidence of the model only based on the generated explanation's confidence. The confidence is most prominently improved for the Movies dataset, where it is also the dataset with the largest improvements for supervised explanation generation with \confidence\ objective. 

\subsection{Unsupervised Rationale Generation}
~\label{sec:unsup}
We explore how well explanations can be generated without supervision from human explanation annotations.
Table~\ref{tab:results:usupervised} shows that the performance of the unsupervised rationales is limited with an up to 47 F1 point decrease for FEVER compared to the supervised model. 
We assume that as our model encodes the whole input together, this leads to a uniform importance of all sentences as they share information through their context. While joint encoding improves the target prediction for complex reasoning datasets especially with more than one explanation sentence, this also limits the unsupervised learning potential of our architecture. As the model is not supervised to select explanations close to human ones, improving the diagnostic properties has a limited effect in improving the results w.r.t. human annotations.

\begin{table}[t]
\centering
\fontsize{10}{10}\selectfont
\begin{tabular}{lrrr}
\toprule
\textbf{Method} & \textbf{FEVER} & \textbf{MultiRC} & \textbf{Movies}\\ \midrule
Sup. & 93.9 {\scriptsize $\pm$0.1} & 78.3 {\scriptsize $\pm$0.1} & 73.8 {\scriptsize $\pm$0.5}\\
UnS.   & 56.1 {\scriptsize $\pm$0.4}& 34.8 {\scriptsize $\pm$7.6}& 50.0 {\scriptsize $\pm$1.8}\\ 
UnS.+DC & 46.9 {\scriptsize $\pm$0.4}& 38.1 {\scriptsize $\pm$3.2} & 63.8 {\scriptsize $\pm$1.2}\\
UnS.+F & 51.6 {\scriptsize $\pm$0.3}& 24.4 {\scriptsize $\pm$5.2} & 64.6 {\scriptsize $\pm$0.4}\\
UnS.+CI & 57.5 {\scriptsize $\pm$0.4}& 25.4 {\scriptsize $\pm$3.4}& 60.0 {\scriptsize $\pm$1.6}\\ 
UnS.+ALL & 57.3 {\scriptsize $\pm$0.2}& 37.4 {\scriptsize $\pm$6.4}& 63.6 {\scriptsize $\pm$0.3}\\ 
\bottomrule
\end{tabular}
\caption{Performance on the explanation generation task without human annotation supervision (UnS.).}
\label{tab:results:usupervised}
\end{table}

\section{Discussion}
\label{sec:discussion}
\subsection{Question/Claim Only Bias}
\label{sec:bias}
Prior work has found that models can learn spurious correlations between the target task and portions of the input text, e.g., predicting solely based on the claim to be fact checked ~\cite{schuster-etal-2019-towards}, regardless of the provided evidence. In our experiments, the input for FEVER and MultiRC also contains two parts - a claim or a question-answer pair and evidence text, where the correct prediction of the target always depends on the evidence.  
Suppose the models do not consider the second part of the input when predicting the target task. In that case, efforts to improve the generated explanations will not affect the target task prediction as it does not rely on that part of the input. 

Table~\ref{tab:results:adversarial} shows target task performance of models trained on the whole input, but using only the first part of the input at test time. We find that, given the limited input, the performance is still considerable compared to a random prediction. For FEVER, the performance drops only with 14 F1-score points to 75.6 F1-score. This could explain the small relative improvements for FEVER when including diagnostic properties as training objectives, where the prediction does not rely on the explanation to a large extent.

Another interesting finding is that including diagnostic properties as training objectives decreases models' performance when a supporting document is not provided. We assume this indicates the properties guide the model to rely more on information in the document than to learn spurious correlations between the question/claim and the target only. The \consistency\ and \confidence\ property lead to the largest decrease in model's performance on the limited input. This points to two potent objectives for reducing spurious correlations.

\begin{table}[t]
\centering
\fontsize{10}{10}\selectfont
\begin{tabular}{llll}
\toprule
\textbf{Dataset} & \textbf{Method} & \textbf{F1-C} & \textbf{Acc-C}\\
\midrule
\multirow{7}{*}{\bf FEVER} & Random & 26.1 {\scriptsize $\pm$4.3}& 37.1 {\scriptsize $\pm$5.6}\\
& Sup. & 75.6 {\scriptsize $\pm$0.3} & 75.7 {\scriptsize $\pm$0.3} \\ 
& Sup.+DC & 68.2 {\scriptsize $\pm$0.2} & 75.6 {\scriptsize $\pm$0.3} \\
& Sup.+F & 73.4 {\scriptsize $\pm$0.4} & 73.9 {\scriptsize $\pm$0.3} \\
& Sup.+CI & 73.2 {\scriptsize $\pm$0.4} & 73.7 {\scriptsize $\pm$0.4}\\ 
& Sup.+ALL &  73.5 {\scriptsize $\pm$0.2} & 73.8 {\scriptsize $\pm$0.4} \\
& Sup. on whole input & 89.3  {\scriptsize $\pm$0.4}& 89.4  {\scriptsize $\pm$0.3}\\ 
\midrule
\multirow{7}{*}{\bf MultiRC} & Random & 26.1 {\scriptsize $\pm$5.5} & 31.6 {\scriptsize $\pm$5.9}\\
& Sup. & 59.4 {\scriptsize $\pm$0.8} & 63.5 {\scriptsize $\pm$0.9} \\
& Sup.+DC & 54.5 {\scriptsize $\pm$0.9}& 61.3 {\scriptsize $\pm$1.2}\\
& Sup.+F & 57.8 {\scriptsize $\pm$0.8} & 61.4 {\scriptsize $\pm$0.6}\\
& Sup.+CI & 49.7 {\scriptsize $\pm$0.8}& 60.1 {\scriptsize $\pm$0.2}\\
& Sup.+ALL & 59.0 {\scriptsize $\pm$0.3} & 61.0 {\scriptsize $\pm$0.2}\\
& Sup. on whole input & 71.0 {\scriptsize $\pm$0.3}& 71.4 {\scriptsize $\pm$0.3}\\ 
\bottomrule
\end{tabular}
\caption{Performance of the models for the downstream task when provided with the query-answer part only.}
\label{tab:results:adversarial}
\end{table}

\subsection{Explanation Examples}
\begin{table}[t]
\small
\centering
\begin{tabular}{p{225pt}}
\toprule
\textbf{Question:} What colors are definitely used in the picture Lucy drew?; \textbf{Answer:} Yellow and purple; \textbf{Label:} True\\
\textbf{Predicted: Sup} True, p=.98; \textbf{Sup+DC} True, p=.99 \\
\textbf{E-Sup:} She draws a picture of her family. She makes sure to draw her mom named Martha wearing a purple dress, because that is her favorite. She draws many yellow feathers for her pet bird named Andy.\\
\textbf{E-Sup+S:} She makes sure to draw her mom named Martha wearing a purple dress, because that is her favorite. She draws many yellow feathers for her pet bird named Andy.\\
\midrule
\textbf{Claim:} Zoey Deutch did not portray Rosemarie Hathaway in Vampire Academy.; \textbf{Label:} REFUTE\\
\textbf{Predicted: Sup} refute, p=.99; \textbf{Sup+F} refute, p=.99\\
\textbf{E-Sup:} Zoey Francis Thompson Deutch (born November 10, 1994) is an American actress. \\
\textbf{E-Sup+F:} She is known for portraying Rosemarie ``Rose'' Hathaway in Vampire Academy(2014), Beverly in the Richard Link later film Everybody Wants Some!! \\ \midrule
\textbf{E-S/S+CI}: For me, they calibrated my creativity as a child; they are masterful, original works of art that mix moving stories with what were astonishing special effects at the time (and they still hold up pretty well).; \textbf{Label:} Positive\\
\textbf{Predicted: Sup} negative, p=.99 \textbf{Sup+CI} positive, p=.99\\

\bottomrule
\end{tabular}
\caption{Example explanation predictions that are changed by including the diagnostic properties as training objectives for explanation generation.}

\label{tab:examples}
\end{table}

Table~\ref{tab:examples} illustrates common effects of the diagnostic properties. We find \consistency\ commonly improves explanations by removing sentences unrelated to the target prediction, as in the first example from MultiRC. This is particularly useful for MultiRC, which has multiple gold explanation sentences. For FEVER and Movies, where one  sentence is needed, the property brings smaller improvements w.r.t. human explanation annotations. 

The second example from FEVER illustrates the effect of including \faithfulness\ as an objective. Naturally, for instances classified correctly by the supervised model, their generated explanation is improved to reflect the rationale used to predict the target. However, when the prediction is incorrect, the effect of the \faithfulness\ property is limited.

Finally, we find \confidence\ often re-calibrates the prediction probabilities of generated explanations and predicted target tasks, which does not change many target predictions. This explains its limited effect as an additional training objective. The re-calibration also influences downstream task prediction confidence, as in the last example from the Movies dataset. This is a side effect of optimising the property while training the target task
, where both explanation and target prediction confidence can be changed to achieve better alignment.

\section{Conclusion}
In this paper, we study the use of diagnostic properties for improving the quality of generated explanations. 
We find that including them as additional training objectives improves downstream task performance and generated explanations w.r.t. human rationale annotations. Moreover, using only the diagnostic properties as training objectives does not lead to a good performance compared to only using human rationale annotations. The latter indicates the need for human rationale annotations for supervising a model to base its predictions on the correct rationales. In future, we plan to experiment with application tasks with longer inputs, where current architectures have to be adjusted to make it computationally possible to encode longer inputs.

\section*{Acknowledgments}

$\begin{array}{l}\includegraphics[width=1cm]{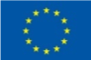} \end{array}$ The research documented in this paper has received funding from the European Union's Horizon 2020 research and innovation programme under the Marie Sk\l{}odowska-Curie grant agreement No 801199. Isabelle Augenstein's research is further partially funded by a DFF Sapere Aude research leader grant.

\bibliography{anthology,acl2021}

\clearpage
\appendix

\section{Datasets}
In our experiments, we employ the FEVER, MultiRC, and the Movies datasets, which have explanation annotations along the annotations for the target task. Table~\ref{tab:datasets} provides more information about the used datasets.
\begin{table}[t]
\centering
\small
\begin{tabular}{lp{60pt}rr}
\toprule
\textbf{Dataset} & \textbf{Size} & \textbf{Doc.Len} & \textbf{Rationale \%} \\ \midrule
FEVER & Train 97957, \newline Val 6122, \newline Test 6111 & 326 & 20.0\\
MultiRC & Train 24029, \newline Val 3214, \newline Test 4848 & 302 & 17.4\\
Movies & Train 1600, \newline Val 150, \newline Test 200 & 774 & 9.35\\
\bottomrule
\end{tabular}
\caption{Dataset statistics.}
\label{tab:datasets}
\end{table}
In the ERASER benchmark the FEVER dataset is simplified compared to the original one -- only claims that require one document for prediction are preserved, and all claims with a label 'Not enough info' are discarded as well.

\section{Experimental Setup}
The model's hyper-parameters are fine-tuned with a grid-search on the validation splits: learning rate of 2e-5 from the values \{1e-5, 2e-5, 3e-5, 4e-5, 5e-5\}; sparsity level of $\lambda=0.5$ from values \{0.3, 0.5, 0.7\}; 10/10/1 words to mask for the \consistency\ property for MultiRC/FEVER/Movies from values \{2, 5, 10\}). We train the models with a an Adam~\cite{DBLP:journals/corr/KingmaB14} optimiser for up to 10 epochs, with a batch size of 8, and select the checkpoint with the best validation performance. For Movies, we use a window with size 500 and stride 300 to encode text of length up to 1200 tokens (chosen from values \{1000, 1200, 1400\}), as the input length exceeds BERT's limit. For FEVER and Movies, we take the sentence with the highest score as there are multiple explanation annotations for one instance, and the model might select more than one. This also allows for a direct comparison with the models of \citet{glockner-etal-2020-think,deyoung-etal-2020-evidence} on these datasets.

All experiments were run on a single NVIDIA TitanX GPU with 24GB memory and 4 Intel Xeon Silver 4110 CPUs. Only the experiments for the Movies dataset were performed on two TitanX GPUs, where due to the long input text, more computational power is required. Training the models for the Movies dataset took $\sim 2$ hours, for the FEVER dataset $\sim 7$ hours, and for MultiRC ~$\sim 3$ hours.

The models were evaluated with accuracy, precision, recall, F1 score, with implementation from scikit. We choose the measures for each task following related work~\cite{glockner-etal-2020-think}. We evaluate the measures over three models and report the mean and the standard deviation for the measures. \url{https://scikit-learn.org/stable/modules/generated/sklearn.metrics.precision_recall_fscore_support.html} and is defined as follows:
\begin{equation*}
  Accuracy (Acc) = \frac{\mathrm{TP+FP}}{\mathrm{TP} + \mathrm{FP}+\mathrm{TN}+\mathrm{FN}} 
\end{equation*}
\begin{equation*}
  Precision (P) = \frac{\mathrm{TP}}{\mathrm{TP} + \mathrm{FP}} 
\end{equation*}
\begin{equation*}
  Recall (R) = \frac{\mathrm{TP}}{\mathrm{TP} + \mathrm{FN}} 
\end{equation*}
\begin{equation*}
  F1 = \frac{2*\mathrm{P}*\mathrm{R}}{\mathrm{P}+\mathrm{R}} 
\end{equation*}
where TP is the number of true positives, FP is the number of false positives, and FN is the number of false negatives.  We compute macro-average F1 scores for target and explanation prediction. For macro-average F1 score, we compute the metric for each class and then take the average, which treats all classes equally.

For the MultiRC and the FEVER datasets, where there are multiple possible explanation annotations per instance, we select the one that maximises the overlap between the gold and the predicted explanation sentences.

\subsection{Human Rationale Annotations}
We note that, in knowledge-intensive tasks like fact checking and question answering, human rationales point to regions in the text containing the information needed for prediction. 
Identifying the required information becomes an important preliminary for the correct prediction rather than a plausibility indicator~\cite{jacovi-goldberg-2020-towards}, and is evaluated as well (e.g., FEVER score). 
Moreover, when multiple sentence sets contain a sufficient rationale (ERASER benchmark provides comprehensive test sets), we consider predicting any of them to be a correct explanation prediction. 
Finally, as we are interested in predictions based on the correct predicted explanation, we measure the joint performance on these tasks, following \citet{glockner-etal-2020-think} and FEVER task evaluation. 
We also note that, as explanations for knowledge-intensive tasks contain the information needed for making the prediction as annotated by human annotators, we assume they are aligned with human expectations~\cite{jacovi2020aligning}, which is, however, not the intended goal of this paper. With the Faithfulness objective, we optimise for causal alignment.

\end{document}